\documentclass[conference]{IEEEtran}
\IEEEoverridecommandlockouts
\usepackage{cite}
\usepackage{amsmath,amssymb,amsfonts}
\usepackage{algorithm}
\usepackage{algpseudocode}
\usepackage{graphicx}
\usepackage{booktabs}
\usepackage{multirow}
\usepackage{textcomp}
\usepackage{xcolor}
\usepackage{url}
\usepackage{dblfloatfix}
\graphicspath{{./}}

\def\BibTeX{{\rm B\kern-.05em{\sc i\kern-.025em b}\kern-.08em
    T\kern-.1667em\lower.7ex\hbox{E}\kern-.125emX}}

\begin{document}

\title{Are the High-Weight Neurons the Important Ones in Image Classification Neural Networks?}

\author{\IEEEauthorblockN{Qitao Chen}
\IEEEauthorblockA{\textit{WeBank Institute of FinTech} \\
\textit{Shenzhen University}\\
Shenzhen, China \\
2210515017@email.szu.edu.cn}
\and
\IEEEauthorblockN{Dongfu Yin\thanks{$^{*}$Corresponding author.}$^{*}$}
\IEEEauthorblockA{\textit{Guangdong Laboratory of Artificial} \\
\textit{Intelligence and Digital Economy (SZ)}\\
Shenzhen, China \\
yindongfu@gml.ac.cn}
\and
\IEEEauthorblockN{Xirui Yang}
\IEEEauthorblockA{\textit{Aberdeen Institute of Data Science} \\
\textit{and Artificial Intelligence}\\
\textit{South China Normal University}\\
Foshan, China \\
20233803084@m.scnu.edu.cn}
\and
\IEEEauthorblockN{Zhaoye Li}
\IEEEauthorblockA{\textit{Guangdong Laboratory of Artificial} \\
\textit{Intelligence and Digital Economy (SZ)}\\
Shenzhen, China \\
2695223647@qq.com}
\and
\IEEEauthorblockN{Liang Xiao}
\IEEEauthorblockA{\textit{SF Express Co., Ltd} \\
Shenzhen, China \\
xiaoliang021@sf-express.com}
}

\maketitle

\begin{abstract}
As neural network models for image classification advance, neurons play critical roles in pruning, backdoor defense, and interpretability. Yet existing work lacks clarity on the weight--importance relationship. We address this with an importance assessment method built on three experiments: quantifying the overlap between high-weight and accuracy-impacting units, analyzing the effect of perturbing high-weight units, and testing post-retraining accuracy after high-weight unit ablation. Throughout this work the unit of analysis is the individual weight parameter---a single scalar in the flattened weight tensor---which we refer to as a \emph{weight-level neuron}; our conclusions therefore concern weight magnitude rather than the function of a conventional unit such as a convolutional filter. Experiments on CIFAR-10 and Mini-ImageNet reveal several patterns. Overlap analysis shows that the top 10\% high-weight neurons overlap with the important ones by only about 25\% at most, dropping further in subsequent intervals. Perturbation tests find that top 10\% high-weight neurons cause 45--80\% accuracy degradation under certain operations, compared with 3--7\% for random perturbations, yet about a third of them show minimal impact. Ablation--retraining results show that removing the top 10\% high-weight neurons leaves accuracy 10--20\% below baseline with no recovery; subdividing this region localizes the irrecoverable damage to a narrow band around the top 0.1\%, while ablating the remaining high-weight sub-intervals permits near-full recovery. Some low-weight intervals also show 10--17\% degradation when perturbed, comparable to mid-range high-weight neurons; this non-monotonic behaviour, however, is observed only for particular datasets and perturbation operations and lies close to the random baseline, so we report it as an observation rather than an established effect. The two findings that hold consistently across architectures and datasets are that the overlap between high-weight and important neurons is uniformly low, and that irrecoverable damage concentrates in a very narrow high-weight band. Together they challenge the assumed equivalence between weight magnitude and importance and offer refined insights into neuron roles, suggesting possible directions such as importance-aware protection of critical high-weight parameters and pruning of non-critical ones.
\end{abstract}


\section{Introduction}\label{sec:intro}
As the fundamental building blocks of neural networks, neurons possess unique learning and memory capabilities that serve as the core driving force behind the remarkable performance of these networks. Specifically, the learning capability allows neurons to adjust their connection weights (i.e., synaptic strengths) based on input data and feedback signals, enabling the network to gradually extract discriminative features from complex datasets. The memory capability, on the other hand, enables neurons to retain the learned information over time, ensuring the stability of the network's performance on previously seen tasks. Consequently, research on neurons has proliferated, with applications in areas such as pruning~\cite{Fanxu2020Pruning, Anusha2024ResPrune}, compression~\cite{Cheng2024Survey, Zonglei2024Survey}, and adversarial attacks~\cite{Abomakhelb2025Review, Hubert2024Adversarial, yadulla2025lightweight}.

Traditional approaches often consider neurons with large absolute values to be the most important, particularly in the context of neural network pruning, where neurons with smaller absolute values are typically pruned. Recent weight-based pruning methods focus on comparing the importance of filters or generated feature maps within specific convolutional layers, thereby retaining critical components and removing non-essential ones. For example, the absolute values of the weights of each convolutional kernel are used as a criterion, with kernels of smaller absolute value being pruned~\cite{Li2017Pruning}; the importance of feature maps is ranked according to weight~\cite{Mao2021Pruning}; redundant flattened-feature dimensions are identified and pruned post hoc according to their contribution to classification performance~\cite{Hu2017NetworkTrimming}; and feature-map channel importance is assessed using the scaling factor in batch-normalization layers~\cite{Liu2017Learning}. However, Morcos et al.~\cite{morcos2018importance} proposed a neuroscience-inspired methodology that evaluates the significance of neuronal groups in deep neural networks through microscopic analysis of neuron ablation effects. Their findings demonstrate that interpretable single neurons exhibit no greater functional importance than their less interpretable counterparts. While these explainable units offer partial intuitive understanding, their contribution to overall network performance proves statistically insignificant. This necessitates new evaluation frameworks that objectively quantify importance through quantitative metrics rather than subjective interpretability assessments.

As bridges connecting microscopic mechanisms with macroscopic behaviors, interpreting neuronal population-level coordination and feature encoding facilitates a deeper understanding of neural network operations, particularly how local feature combinations enable emergent abstract semantic comprehension. Investigating correlations between neuronal weights and functional significance establishes cross-scale associations: integrating granular neuron-level analysis with system-wide emergent properties could achieve breakthroughs in understanding interactions spanning from local component interactions to global system behaviors.

Before proceeding, we make the granularity of our analysis explicit, because two different notions of ``neuron'' are current in this literature. A \emph{conventional neuron} is a whole computational unit---a convolutional filter or a fully connected unit---and comprises many weights; a \emph{weight-level neuron}, the unit of analysis adopted here, is a single scalar in the flattened weight tensor, i.e.\ one connection. Ablating a weight-level neuron sets one scalar to zero and removes a single connection, which is not the same operation as deactivating an entire unit, as studied by Morcos et al.~\cite{morcos2018importance}, Zhou et al.~\cite{Zhou2018Revisiting}, and Bau et al.~\cite{Bau2020Understanding}. Our findings therefore speak to the relationship between weight magnitude and importance, and unit-level ablation remains a complementary line of work. To thoroughly evaluate the correlation between weight-level neuron weights and their significance within the model, we designed and implemented the following three experiments.

\begin{itemize}
\item \textit{Analyzing the overlap between high-weight neurons and key accuracy contributors:} we quantified and compared the overlap between neurons with high weights and those experimentally validated as significantly impacting model prediction accuracy.
\item \textit{Evaluating model robustness under perturbation strategies:} perturbations were applied to high-weight neurons, and the corresponding decline in model accuracy was observed and recorded. This helped us assess the contribution of these high-weight neurons to model accuracy and stability.
\item \textit{Retraining and performance evaluation after removing neurons:} neurons from different weight intervals within the network were removed, and the remaining parts were retrained to further investigate the intrinsic relationship between weight and neuron importance.
\end{itemize}

These three experiments do not measure ``importance'' in a single sense; each operationalizes it differently---as set overlap with an accuracy-impact ranking, as sensitivity to perturbation, and as post-retraining irreplaceability---and the three proxies are complementary rather than interchangeable. We therefore report them separately and do not treat a neuron flagged by one as necessarily important under the others; the term ``importance'' should be read as shorthand for this trio of proxies rather than as a single homogeneous quantity.

By adopting a holistic perspective to evaluate the importance of all neurons in the network, rather than limiting the analysis to a subset or a specific type of neuron, we achieve a more efficient and comprehensive assessment. We conducted experiments on the CIFAR-10 and Mini-ImageNet benchmark datasets. Contrary to the traditional perspective, our experimental results demonstrate that not all neurons within the top 10\% of weights can be categorized as important, while neurons in the bottom 10\% of weights are also capable of exerting a significant impact on model performance. This finding aids in deepening our understanding of the internal workings of neural networks, improving methods for analyzing and debugging them, advancing interpretability research, guiding the design and optimization of neural networks, and enhancing the recognition of neural network robustness.

\section{Related Work}\label{sec:related}
Evaluating the importance of neurons has long been a pivotal topic in neural network research~\cite{Yongqi2025Towards, blakeman2025nemotron}. In this section, we review studies from the fields of interpretability and pruning, with a particular focus on the methodologies that have been employed to select and assess the importance of neurons.

\subsection{Identification of Key Neurons for Model Interpretability}\label{subsec:interp}
The interpretability of neurons has been a widely studied area in recent years~\cite{Antamis2024Interpretability, Dalal2023Understanding, Younis2024MTS2Graph}. Neurons exhibit capabilities such as nonlinear learning and information storage. Visualizing neuron activations within the network facilitates an understanding of how the model responds to different inputs and processes information. Bau et al.~\cite{Bau2017Network, Bau2018Gan, Bau2020Understanding} systematically identified the semantics of neurons in image classification and generation networks by constructing a pixel-level dataset with semantic concepts and introducing the network dissection framework. The importance of neuron feature representation was quantified by evaluating the consistency between hidden-layer neurons and semantic concepts. Nguyen et al.~\cite{Nguyen2016Plug} and Olah et al.~\cite{Olah2017Featurevisualization} expanded beyond the construction of semantic datasets by optimizing inputs through synthetic features to activate specific neurons. After identifying particular features corresponding to individual neurons, Hase et al.~\cite{Hase2024Does} localized specific facts to individual neurons and modified the storage of these facts by editing the neurons responsible for this knowledge.

Previous studies have employed techniques such as weight masking~\cite{khanna2023differentiable}, feature activation analysis~\cite{Wang2025NeuronPath}, and ablation experiments~\cite{Li2024OptimalAblation} to identify neurons that are critical for various tasks. However, these investigations have not explored the relationship between neurons and their corresponding weights. Conducting in-depth research on whether neurons with larger absolute values constitute a key neuron group is essential for identifying those that play a dominant role in the network's decision-making process. This not only helps to uncover the intricate information-processing mechanisms within neural networks but also enhances the transparency and reliability of model predictions.

\subsection{Selection of Critical Neurons in Model Pruning}\label{subsec:pruning}
Neural network pruning can be categorized based on granularity into unstructured pruning~\cite{Chen2025DLP, Wimmer2022Interspace, Frantar2023Sparsegpt} and structured pruning~\cite{Gongfan2023DepGraph, Ma2023LLMPruner}. Unstructured pruning is particularly effective for reducing the memory footprint of neural networks during both training and inference. Notably, neuron-based pruning facilitates a more flexible and tailored final architecture. Among neuron-based methods, Han et al.~\cite{Song2015Learning} proposed an unstructured pruning technique that identifies neurons with absolute weight values below an empirical threshold as unimportant; these neurons are pruned to achieve a sparse weight pattern. Similarly, Sun et al.~\cite{Sun2023Simple} pruned non-essential neurons by evaluating weights with the smallest magnitudes, multiplied by the corresponding input activations for each output.

In contrast, structured pruning adopts a hardware-friendly approach by removing convolutional kernels based on importance rankings or similarity. For instance, Basha et al.~\cite{Shabbeer2024Deep} assess filter (a combination of neurons) importance during iterative pruning by computing the similarity between filters. Yu et al.~\cite{Yu2023Recursive} combined the inverse input-autocorrelation matrix with the weight matrix to evaluate and prune unimportant input channels or nodes in each CNN layer, enabling the removal of redundant filters while maintaining a low error rate. Zheng et al.~\cite{Yongbin2024Novel} calculate the importance scores of filters based on their weights, subsequently removing those with low scores while considering both their direct and indirect impacts. By eliminating redundant or less contributive neurons, computational cost and storage requirements can be reduced while preserving model performance. The prevailing practice has been to prune neurons with smaller weights~\cite{Gong2024Lightweight, Song2015Learning}, but it remains unclear whether neurons with larger weights truly exert a significant influence; previous studies have not provided relevant research or conclusions on this matter.

\subsection{Positioning of This Work}\label{subsec:positioning}
The three studies closest to ours in spirit differ from it in both the granularity at which importance is defined and the proxy used to measure it. Morcos et al.~\cite{morcos2018importance} ablate whole units and single activation directions and use class selectivity as the candidate proxy for importance, concluding that selectivity does not predict a unit's contribution to generalization. Zhou et al.~\cite{Zhou2018Revisiting} likewise ablate individual units, but take the drop in per-class accuracy as the proxy and show that a small number of units dominates the accuracy of particular classes. Hod et al.~\cite{Hod2021Quantifying} move to a still coarser granularity, partitioning a network into clusters of units and using the functional specialization of each cluster as the proxy. All three therefore operate on the \emph{conventional neuron} or on groups of them, and none of them takes weight magnitude as the quantity under test. What this paper adds is a systematic examination of the correspondence between weight magnitude and importance at the granularity of the individual weight: we rank every weight in the network by absolute value, partition the ranking into intervals, and measure---by set overlap, by perturbation, and by ablation followed by retraining---how much of a network's accuracy each interval actually accounts for. The magnitude-based selection rule that unstructured pruning takes for granted thus becomes the object of study rather than the tool.

\begin{figure*}[t]
\centering
\includegraphics[width=0.92\textwidth]{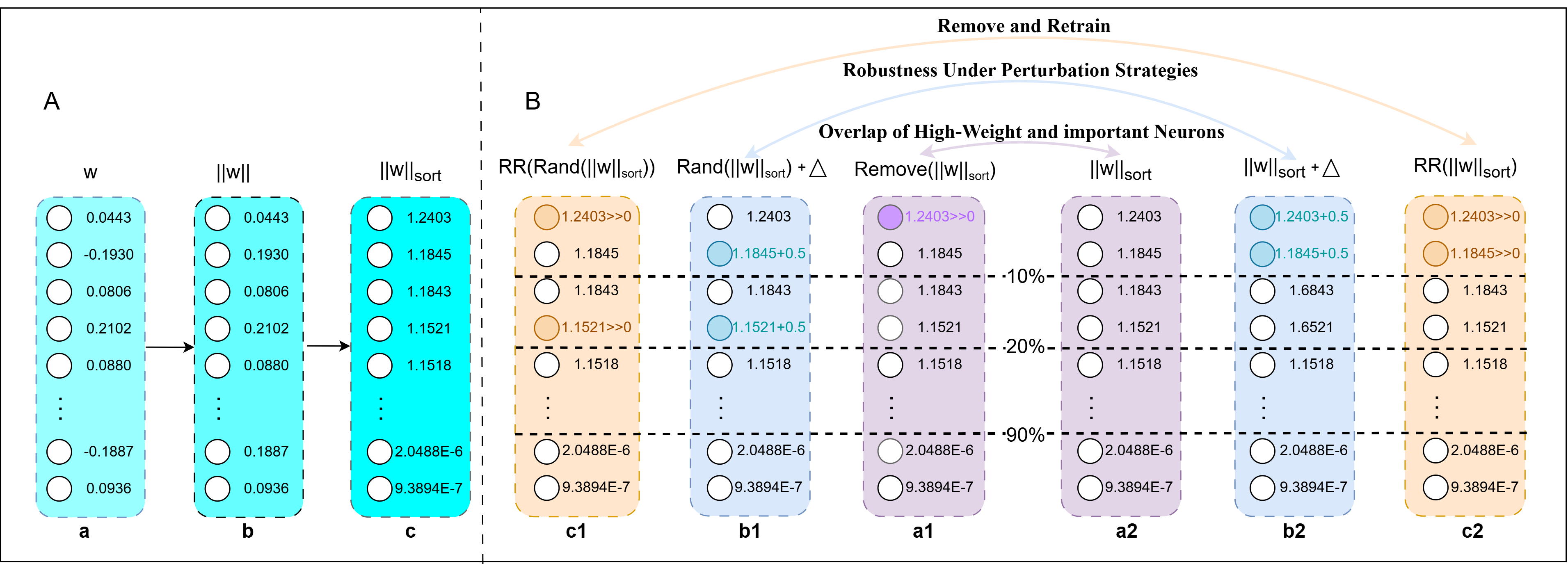}
\caption{Experimental methods and procedures for neuron importance evaluation. Circular nodes represent neurons within the network; colored nodes indicate modified neurons, and the numbers on the edges represent the neuron weights. Panel~(A) illustrates the initial flattened neuron weight matrix, sorted by absolute value: A(a) is the flattened one-dimensional matrix, A(b) denotes the absolute values of all weights, and A(c) is the neuron matrix sorted by absolute value. In panel~(B), dashed lines and their numbers denote the partitioned intervals of neurons. B(a2) is the sorted neuron matrix and B(a1) the ablation of neurons; B(b2) denotes the introduction of perturbations to the weights and B(b1) the random selection of neurons in the same proportion for perturbation; RR in B(c2) signifies retraining of the remaining neurons after ablation, while B(c1) represents random ablation and fine-tuning of the same number of neurons. This figure gives the overall pipeline and the relation between the three experiments and their random-control counterparts; the specific arithmetic of each perturbation operator is not shown here but in Fig.~\ref{fig:strategy}.}
\label{fig:method}
\end{figure*}

\section{Method}\label{sec:method}
It is generally agreed that larger weights typically indicate that the corresponding features are more important or appear more frequently in the input data, whereas smaller weights may correspond to less significant features or even noise. However, despite this common assumption, few studies have thoroughly investigated whether neurons with large weights are truly synonymous with important neurons. Prior research on neuron importance, including the work of Zhou et al.~\cite{Zhou2018Revisiting} and Hod et al.~\cite{Hod2021Quantifying}, has not fully addressed this question. To assess the significance of neurons with large weights, we designed three comprehensive evaluation experiments.

\subsection{Experimental Scheme}
After the neural network is trained, we flatten the pretrained neuron weight matrices from a multidimensional format into a one-dimensional vector, as shown in Fig.~\ref{fig:method}A(a). Subsequently, we apply the absolute-value operation to each weight to eliminate sign information and retain only the magnitude, as shown in Fig.~\ref{fig:method}A(b). Finally, we sort the absolute values to generate an initial neuron weight matrix, as shown in Fig.~\ref{fig:method}A(c). Based on this preprocessing, we designed three experiments, as shown in Fig.~\ref{fig:method}B. As stated in Section~\ref{sec:intro}, the unit of analysis throughout is the weight-level neuron, i.e.\ the individual weight parameter, and every occurrence of ``neuron'' in Sections~\ref{sec:method} and~\ref{sec:exp} refers to such a parameter unless a conventional unit is named explicitly. This weight-level granularity is a deliberate design choice, and its implications are discussed in Section~\ref{sec:exp}.

As shown in Fig.~\ref{fig:method}B(a1) and~(a2), we quantified and compared the overlap between high-weight neurons and those experimentally validated as having a significant impact on model prediction accuracy, in order to analyze whether there is a consistency between high-weight neurons and key accuracy contributors. As illustrated in Fig.~\ref{fig:method}B(b1) and~(b2), to assess the contribution of neurons with different weights to model accuracy and stability, we conducted a neuron perturbation experiment: perturbations were introduced to neurons, and the resulting decline in model accuracy was monitored and recorded. As depicted in Fig.~\ref{fig:method}B(c1) and~(c2), we performed a neuron removal and retraining experiment: neurons from different weight intervals were removed and the remaining neurons were retrained, aiming to further uncover the relationship between weight magnitude and neuron importance.

For reference, Table~\ref{tab:notation} collects the symbols used in Sections~\ref{sec:method} and~\ref{sec:exp}, together with the range of each derived quantity.

\begin{table}[t]
\centering
\caption{Notation used throughout Sections~\ref{sec:method} and~\ref{sec:exp}.}
\label{tab:notation}
\resizebox{\columnwidth}{!}{%
\begin{tabular}{@{}ll l@{}}
\toprule
Symbol & Meaning & Range \\
\midrule
$w,\ b$ & trainable weights and biases & --- \\
$w^{n}_{k,ij}$ & weight at row $i$, column $j$ of filter $k$ in layer $n$ & --- \\
$N$ & total number of weight-level neurons & --- \\
$n$ & number of weight intervals ($n=10$ unless stated) & --- \\
$N_m$ & cumulative \emph{count} of neurons in the first $m$ intervals & $1..N$ \\
$S_{w,m}$ & top-$N_m$ set ranked by $\lvert w\rvert$ & --- \\
$S_{a,m}$ & top-$N_m$ set ranked by accuracy impact & --- \\
$R_{a,w,m}$ & overlap of $S_a$ and $S_w$ within interval $m$, \eqref{eq:overlap} & $[0,1]$ \\
$\Delta$ & perturbation magnitude & --- \\
$\circledast$ & operator of a perturbation strategy ($+,-,\times,\div$) & --- \\
$p_i$ & test accuracy after perturbing interval $i$ & $[0,1]$ \\
$E(p_i)$ & perturbation effect of interval $i$, \eqref{eq:effect} & $[\epsilon,1]$ \\
$R(p_i\mid p)$ & normalized perturbation effect, \eqref{eq:effectratio} & $[0,1]$ \\
$r_i$ & post-retraining accuracy after ablating interval $i$ & $[0,1]$ \\
$R(r_i\mid r)$ & irreplaceability of interval $i$, \eqref{eq:irr} & $[0,1]$ \\
$\epsilon$ & positive constant ensuring non-zero denominators & $10^{-6}$ \\
$m$ (mask) & binary mask freezing ablated parameters & $\{0,1\}^{|w|}$ \\
$T$ & number of retraining epochs & --- \\
\bottomrule
\end{tabular}%
}
\end{table}

\subsection{Overlap Between High-Weight and Key Accuracy Neurons}
In a trained neural network, we adopt a systematic approach to identify critical neurons by iteratively ablating each neuron within the network's architecture. Specifically, for each layer of the weight matrix, we conduct a granular analysis by sequentially setting the weights of individual neurons to zero. This effectively simulates the removal of a single neuron while maintaining the integrity of the network's structure and the parameters of other neurons. Following each ablation, we evaluate the modified network on an independent test dataset. By measuring accuracy, we quantify the impact of each neuron's removal on the model's predictive capability. Through comprehensive comparison of the model's performance before and after each ablation, we pinpoint neurons that exert a substantial influence on the overall output; these neurons are deemed critical, as their absence leads to significant degradation in accuracy.

After training, we take the absolute values of the weights and sort them in descending order, selecting the top-$n$ neurons as the weight-magnitude top-$n$ set $S_w$. We perform ablation on individual neurons and evaluate the network's accuracy after each neuron is ablated. The neurons with the highest impact on accuracy define the accuracy-impact top-$n$ set $S_a$. Finally, we compute the overlap between these two sets, measuring the proportion of neurons in $S_a$ that also appear in $S_w$. An importance-quantification metric is constructed to reflect the relationship between neuron weight magnitude and functional impact:
\begin{equation}
R_{a,w,m} = \frac{\lvert S_{a,m} \cap S_{w,m}\rvert - \lvert S_{a,m-1} \cap S_{w,m-1}\rvert}{N_m - N_{m-1}},
\label{eq:overlap}
\end{equation}
where $m$ indexes the interval and $N_m$ is the cumulative \emph{number} of neurons contained in the first $m$ intervals, so that $N_m - N_{m-1}$ is the number of neurons in the $m$-th interval and numerator and denominator are both counts. $S_{a,m}$ and $S_{w,m}$ denote the accuracy-impact and weight-magnitude top-$N_m$ sets, and $R_{a,w,m}$ is the overlap between $S_a$ and $S_w$ within the $m$-th interval. Because these sets are nested by construction ($S_{a,m-1}\subseteq S_{a,m}$ and $S_{w,m-1}\subseteq S_{w,m}$, both being top-$N$ sets of the same ranking), the numerator counts the neurons that enter the intersection at interval $m$ and is therefore non-negative; it is also bounded above by $N_m-N_{m-1}$, so $R_{a,w,m}\in[0,1]$ and can be read directly as the fraction of the $m$-th interval on which the two rankings agree. A value of $1$ would indicate perfect agreement within the interval and $0$ complete disagreement; the expected value under independent rankings is approximately $N_m/N$, where $N$ is the total number of neurons.

\begin{figure}[t]
\centering
\includegraphics[width=0.95\columnwidth]{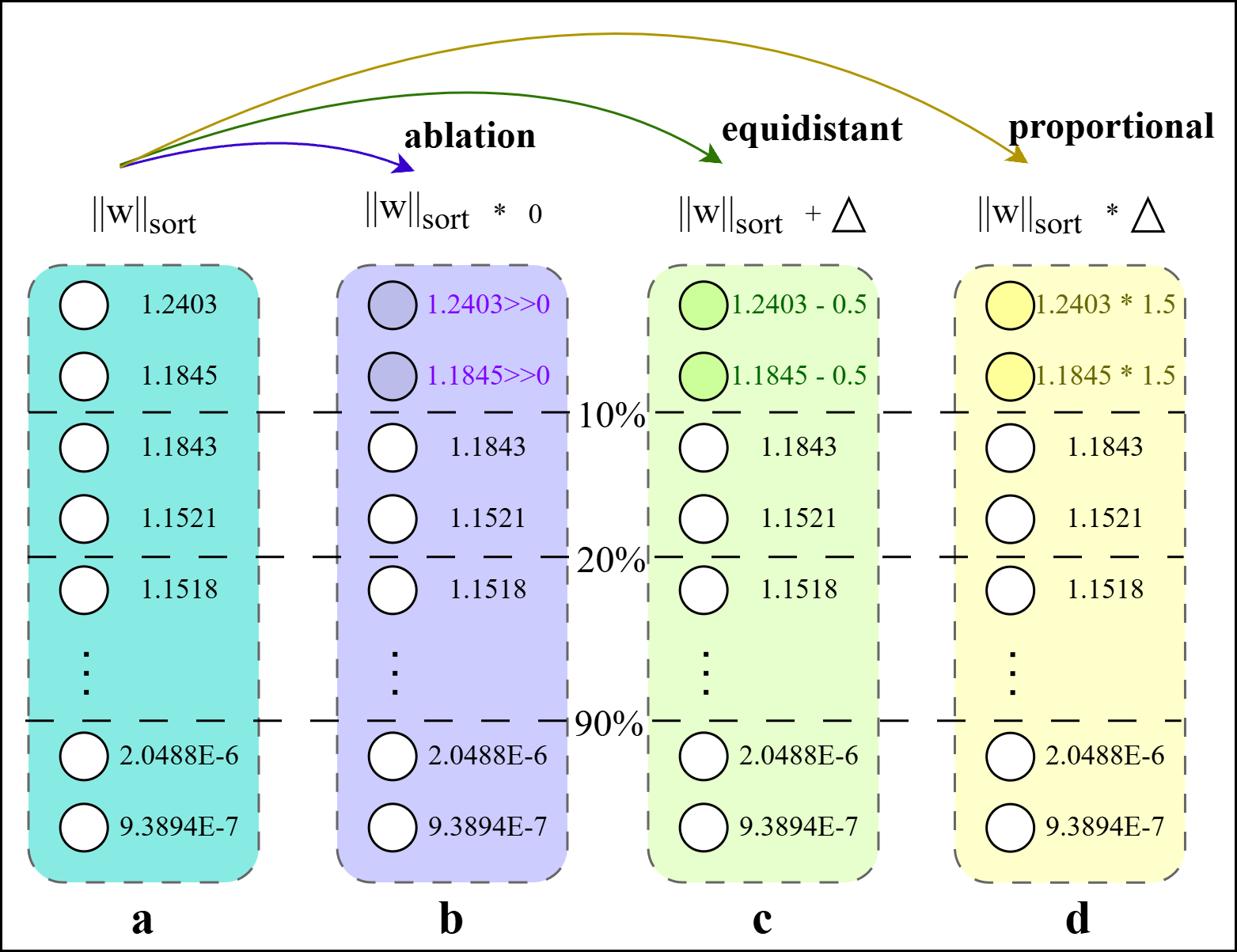}
\caption{Ablation and perturbation strategies for the initial neurons, flattened and sorted by absolute weight value. Complementing the pipeline overview of Fig.~\ref{fig:method}B, this figure specifies how each operator transforms an individual weight value. (a)~is the one-dimensional neuron weight matrix obtained by taking absolute values and sorting in descending order. The purple circle and value in~(b) denote a specified neuron ablation; the green circle and value in~(c) indicate an equidistant perturbation based on a deviation of 0.5; and the yellow circle and value in~(d) denote a proportional perturbation based on a deviation of 0.5.}
\label{fig:strategy}
\end{figure}

\subsection{Model Robustness Under Perturbation Strategies}
To assess the impact of neuron weights on model performance under non-ablation conditions, we design non-ablation perturbations targeting the trained neuron weights. Writing the trainable parameters of a network with $L$ layers by index rather than in block-matrix form,
\begin{equation}
w = \bigl\{\, w^{n}_{k,ij} \,\bigr\},\qquad
b = \bigl\{\, b^{n}_{k} \,\bigr\},
\label{eq:w}
\end{equation}
where $n\in\{1,\dots,L\}$ indexes the layer, $k\in\{1,\dots,k_n\}$ the filter, and $(i,j)$ the position within the filter, so that the convolutional layer $n$ holds $i\,j\,k_n$ weights and $k_n$ biases. Only the weights $w$ are ranked and perturbed: biases are excluded from the ranking, from the interval partition, and from every perturbation, since including a few thousand bias terms whose magnitudes follow a different scale would confound the weight-magnitude ordering under study. Weight-level neurons are ranked by the absolute value of their weights and divided into non-overlapping intervals, ensuring that the relative magnitude of weights within each interval remains consistent.

Three perturbation strategies are then applied: ablation perturbation (Fig.~\ref{fig:strategy}b), equidistant perturbation (Fig.~\ref{fig:strategy}c), and proportional perturbation (Fig.~\ref{fig:strategy}d). In the equidistant strategy, a fixed predefined small value is added to all weights within each interval, assessing the impact of direct numerical changes. In the proportional strategy, weights within each interval are scaled proportionally, exploring how changes in the relative magnitude of weights affect accuracy. Following each perturbation, inference is performed on a separate test set. For a test input $x$, the prediction under a perturbation strategy is
\begin{equation}
\tilde{P}(y\mid x) = \operatorname{softmax}\!\big(F\big((w \circledast \Delta)\ast x + b\big)\big),
\label{eq:perturb}
\end{equation}
where $\circledast$ selects the operator corresponding to a perturbation strategy, $\Delta$ is the perturbation magnitude, $\ast$ is convolution, and $F$ is the network's nonlinear feature extractor. Consistent with the partitioning described above, the bias term $b$ appears unmodified in~\eqref{eq:perturb}: perturbations are applied to weights only. We denote by $p_i$ the classification accuracy on the test set after perturbing the $i$-th interval. Based on this non-ablation perturbation, we define evaluation metrics that relate the weights of neurons in different intervals to their importance:
\begin{align}
E(p_i) &= \max(p_1, p_2, \dots, p_n) + \epsilon - p_i, \label{eq:effect}\\
R(p_i\mid p) &= \frac{E(p_i)}{\sum_{i=1}^{n} E(p_i)}. \label{eq:effectratio}
\end{align}
Here the interval with the highest accuracy after perturbation is taken as the baseline. Adding a small positive constant $\epsilon$ to this baseline and subtracting the perturbed accuracy $p_i$ of each interval yields the perturbation effect $E(p_i)$; the proportion of this effect for the $i$-th interval is $R(p_i\mid p)$. The role of $\epsilon$ is solely to keep $E(p_i)$ strictly positive for the baseline interval itself, so that the denominator of~\eqref{eq:effectratio} cannot vanish. Since $\epsilon$ enters that denominator it does affect the numerical value of $R(p_i\mid p)$, and we therefore fix $\epsilon=10^{-6}$ (in units of accuracy) in all experiments reported here, for both~\eqref{eq:effect}--\eqref{eq:effectratio} and~\eqref{eq:irr}. This value is several orders of magnitude below the smallest accuracy difference we observe, so it leaves the ranking of intervals unchanged while guaranteeing well-defined ratios. Both $R(p_i\mid p)$ and $R(r_i\mid r)$ are non-negative and sum to one over the $n$ intervals.

\subsection{Retraining After Removing High-Weight Neurons}
In contrast to previous approaches, we simultaneously apply neuron pruning and neuron freezing to assess the importance of neurons across different weight intervals (Algorithm~\ref{alg:retrain}). Specifically, for the trained network, the absolute values of the neuron weights are sorted in descending order and the weights of the top-$n$ neurons are set to zero. A mask matrix then ensures that the gradients and weights of these neurons remain unchanged during subsequent retraining, allowing only the unfrozen parameters to be updated. This investigates the contribution of different weight regions to the model's learning capability and their impact on final accuracy.

To further validate the partitioning strategy, we design a comparative experiment in which neurons are randomly selected and the same strategy is applied. This random-freezing strategy simulates an unstructured parameter-reduction method. By comparing the change in accuracy under interval-based zeroing and random freezing, we quantitatively analyze the effect of neurons with different weights on classification performance:
\begin{align}
\tilde{P}(y\mid x) &= \operatorname{softmax}\!\big(F\big(w_{\mathrm{mask}}\ast x + b\big)\big), \label{eq:mask}\\
R(r_i\mid r) &= \frac{r_{\max}+\epsilon-r_i}{n(r_{\max}+\epsilon)-\sum_{i=1}^{n}r_i}. \label{eq:irr}
\end{align}
Here $w_{\mathrm{mask}}$ is the weight matrix in which the top-$n$ neurons are set to zero with non-updatable gradients, $r_i$ is the fine-tuned accuracy after ablating interval $i$, $r_{\max}$ is the maximum such accuracy, $r_{\max}+\epsilon$ is the baseline accuracy, and $n$ is the number of intervals. We refer to $R(r_i\mid r)$ as the neuron \emph{irreplaceability} of interval $i$.

Algorithm~\ref{alg:retrain} keeps two seemingly redundant steps: masking the gradient (line~9) and re-projecting the weights onto the mask after the update (line~11). Both are retained deliberately. Gradient masking alone does not guarantee that ablated entries stay at zero under Adam, because the optimizer maintains first- and second-moment estimates and applies decoupled updates, so a parameter whose gradient is zero at step $t$ can still receive a non-zero update from moments accumulated earlier or from weight decay. The projection in line~11 makes the invariant $w\odot(1-m)=0$ hold exactly at every epoch boundary, independently of the optimizer state.

\begin{algorithm}[t]
\caption{Retraining with top-$n$ ablated neurons}
\label{alg:retrain}
\begin{algorithmic}[1]
\Require original weights $w$, ablation count $n$, epochs $T$
\State $m \gets \mathbf{1}$ \Comment{mask, same shape as $w$}
\State $\mathit{idx} \gets \operatorname{argsort}(\lvert \operatorname{flatten}(w)\rvert,\ \text{descending})$
\State $\mathcal{T} \gets \mathit{idx}[1{:}n]$ \Comment{top-$n$ high-weight neurons}
\ForAll{$i \in \mathcal{T}$}
    \State $w_i \gets 0$;\quad $m_i \gets 0$
\EndFor
\For{$t = 1$ to $T$}
    \State $g \gets \nabla_w \mathcal{L}$ \Comment{gradients on training data}
    \State $g \gets g \odot m$ \Comment{freeze gradients of ablated neurons}
    \State update $w$ using masked gradients $g$
    \State $w \gets w \odot m$ \Comment{re-project; cancels Adam moments}
\EndFor
\State \Return $\operatorname{Evaluate}(w,\ \text{test set})$
\end{algorithmic}
\end{algorithm}

\section{Experiment}\label{sec:exp}
\subsection{Experimental Setup}
Two image classification datasets, CIFAR-10~\cite{Krizhevsky2009Learning} and Mini-ImageNet~\cite{Oriol2023Matching}, were used to train our models. CIFAR-10 contains 60{,}000 color images of size $32\times32$ across 10 classes. Mini-ImageNet contains 60{,}000 color images of size $84\times84$ across 100 classes, with 600 examples per class; its images were reshaped to $224\times224$ pixels and normalized with a mean of 0.5 and a standard deviation of 0.5 before being used as input. For each dataset, the images were partitioned into training (80\%), validation (10\%), and test (10\%) subsets. The experiments were conducted in parallel on four NVIDIA GeForce RTX 3090 GPUs. All models were trained with the Adam optimizer and an initial learning rate of $10^{-4}$; the numbers of training and retraining \emph{epochs} were set to 50 and 20, respectively, the latter corresponding to $T=20$ in Algorithm~\ref{alg:retrain}.

We evaluate three convolutional architectures. LeNet-5~\cite{LeCun1998Gradient} is the basis of the earliest CNN models applied to computer vision; its relatively small number of parameters allows a noticeable impact on accuracy when adjusting individual neurons. To verify that the observed patterns generalize to deeper networks, we additionally adopt ResNet-18 and ResNet-50~\cite{He2016Deep}.

\begin{figure*}[tp]
\centering
\includegraphics[width=\textwidth]{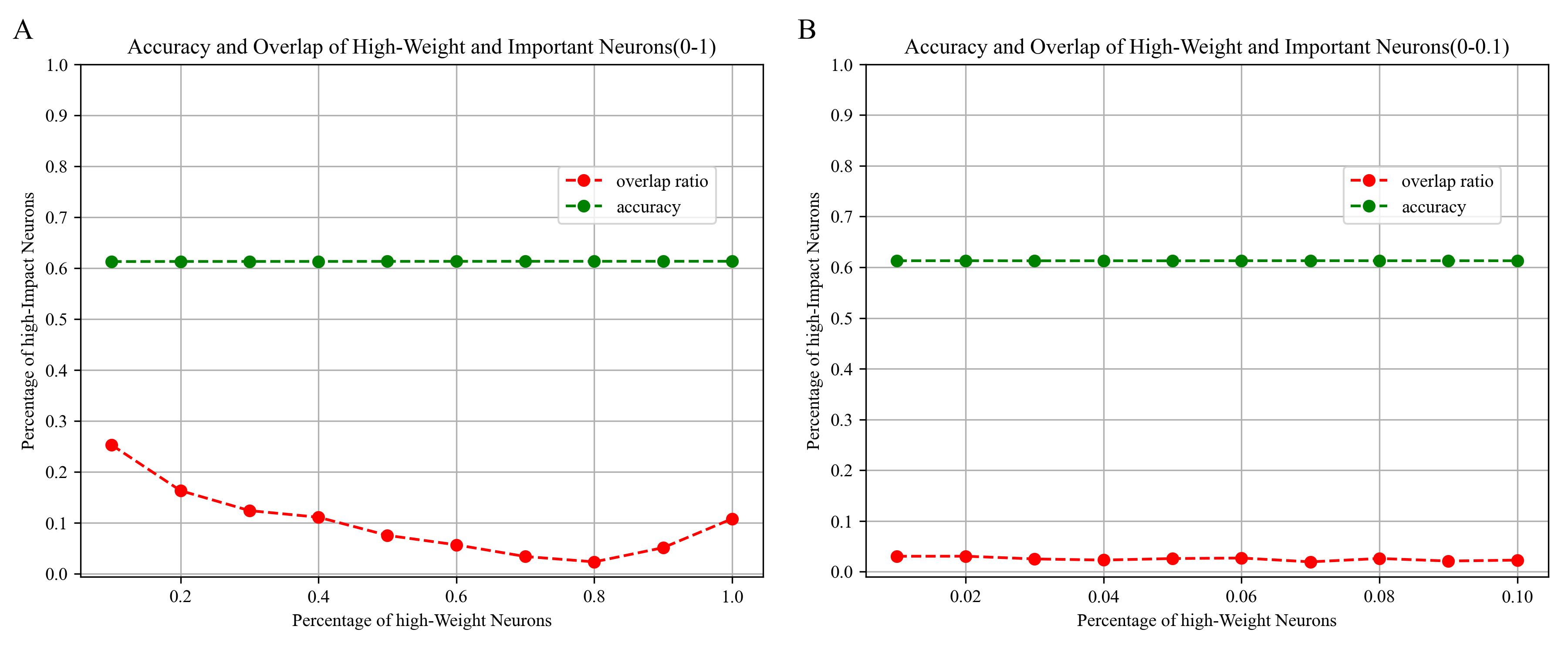}
\includegraphics[width=\textwidth]{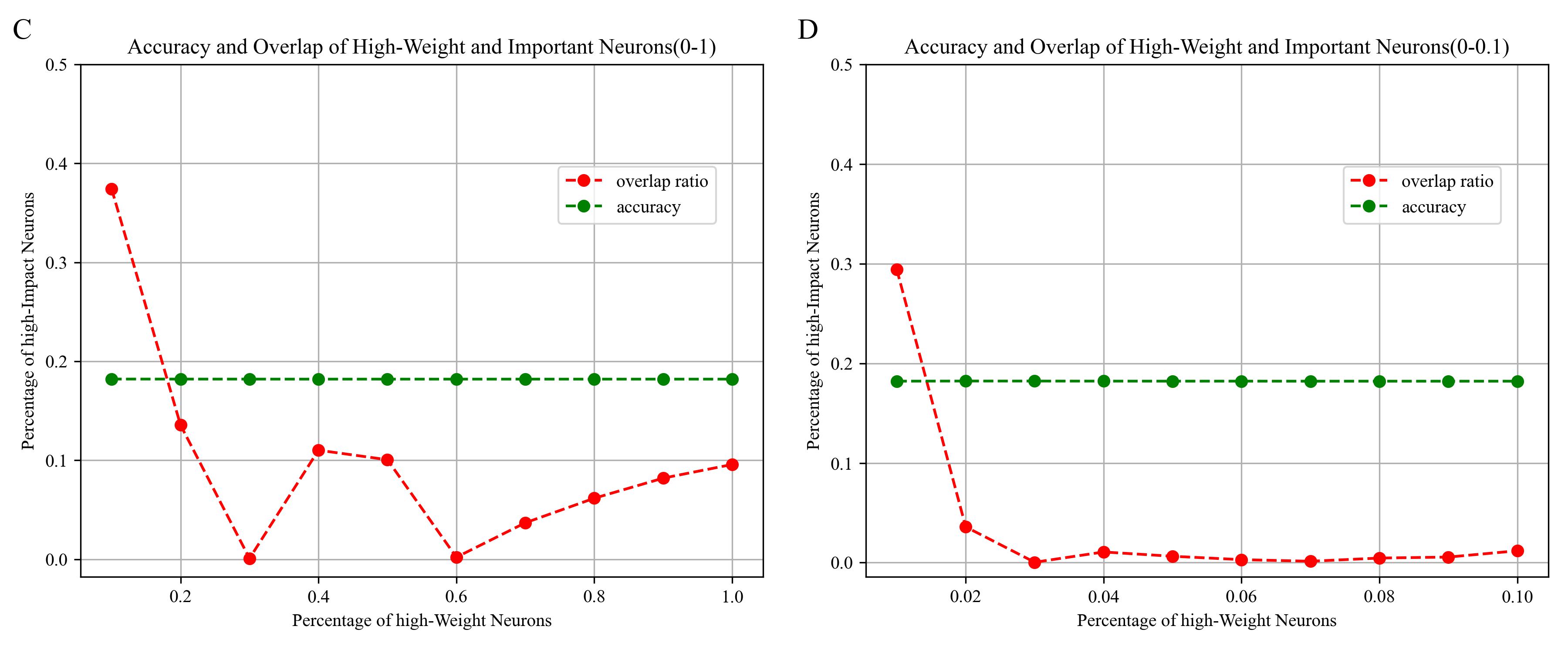}
\caption{Accuracy and overlap between high-weight and important neurons on CIFAR-10 (panels A, B) and Mini-ImageNet (panels C, D). The red line is the proportion of neurons that are both high-weight and important within the same interval (overlap ratio); the green line is the model accuracy reference. Panels A and C span the full range of 10\% intervals; panels B and D zoom into the top 1\%--2\% high-weight neurons.}
\label{fig:overlap}
\end{figure*}

\subsection{Overlap Between High-Weight and Key Accuracy Neurons}
We quantified the overlap and accuracy between weight and importance across non-overlapping 10\% intervals (Fig.~\ref{fig:overlap}). Panels~A and~C explore the overall trend across all intervals. In panel~A (CIFAR-10), even the top 10\% interval, where the overlap is highest, only slightly exceeds 25\% for neurons that are both high-weight and important, indicating a generally low overlap; the red (overlap) curve decreases through the first 80\% of intervals, reaches a minimum, and then rises, while the green accuracy reference stays essentially flat. In panel~C (Mini-ImageNet), the overlap hits local minima around the 30\% and 60\% intervals before rising slightly, following an overall decline-then-rise trend with no interval achieving substantially high values. These fluctuations reinforce that weight alone is not a reliable indicator of importance.

To verify whether more concentrated high-weight ranges yield higher overlap, panels~B and~D focus on the top 1\%--2\% weight neurons. In panel~B (CIFAR-10), the overlap shows no significant advantage, confirming that even narrowly defined high-weight neurons do not strongly correspond to importance. In panel~D (Mini-ImageNet), the very top 1\%--2\% interval correlates more strongly with important-neuron coverage, though subsequent intervals display weak correlation, revealing an importance gradient even among high-weight neurons. Together, these results demonstrate that while the very top 1\%--2\% weight neurons show some correlation with importance, this relationship is limited and inconsistent.

\begin{figure*}[tp]
\centering
\includegraphics[width=0.48\textwidth]{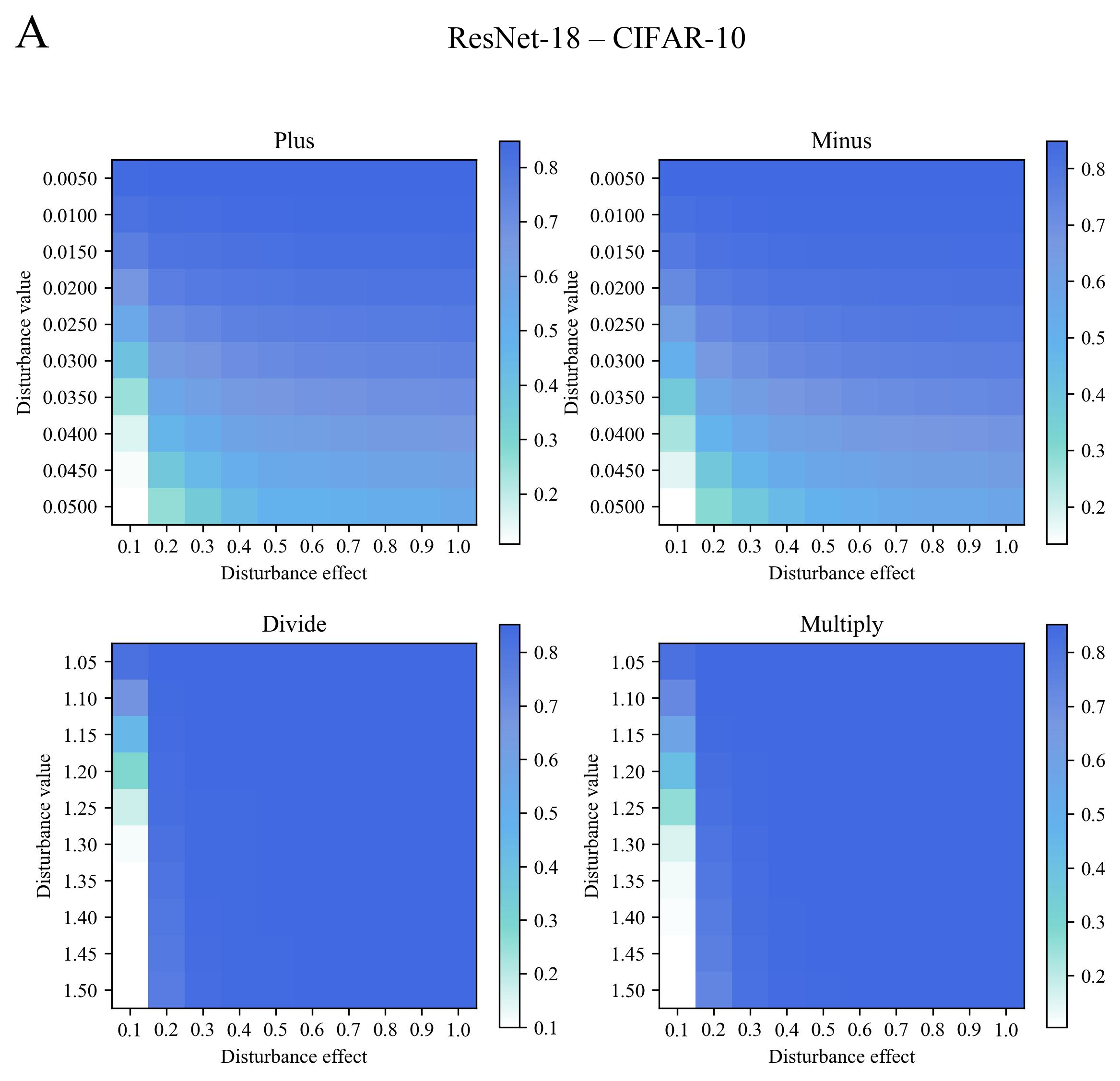}\hfill
\includegraphics[width=0.48\textwidth]{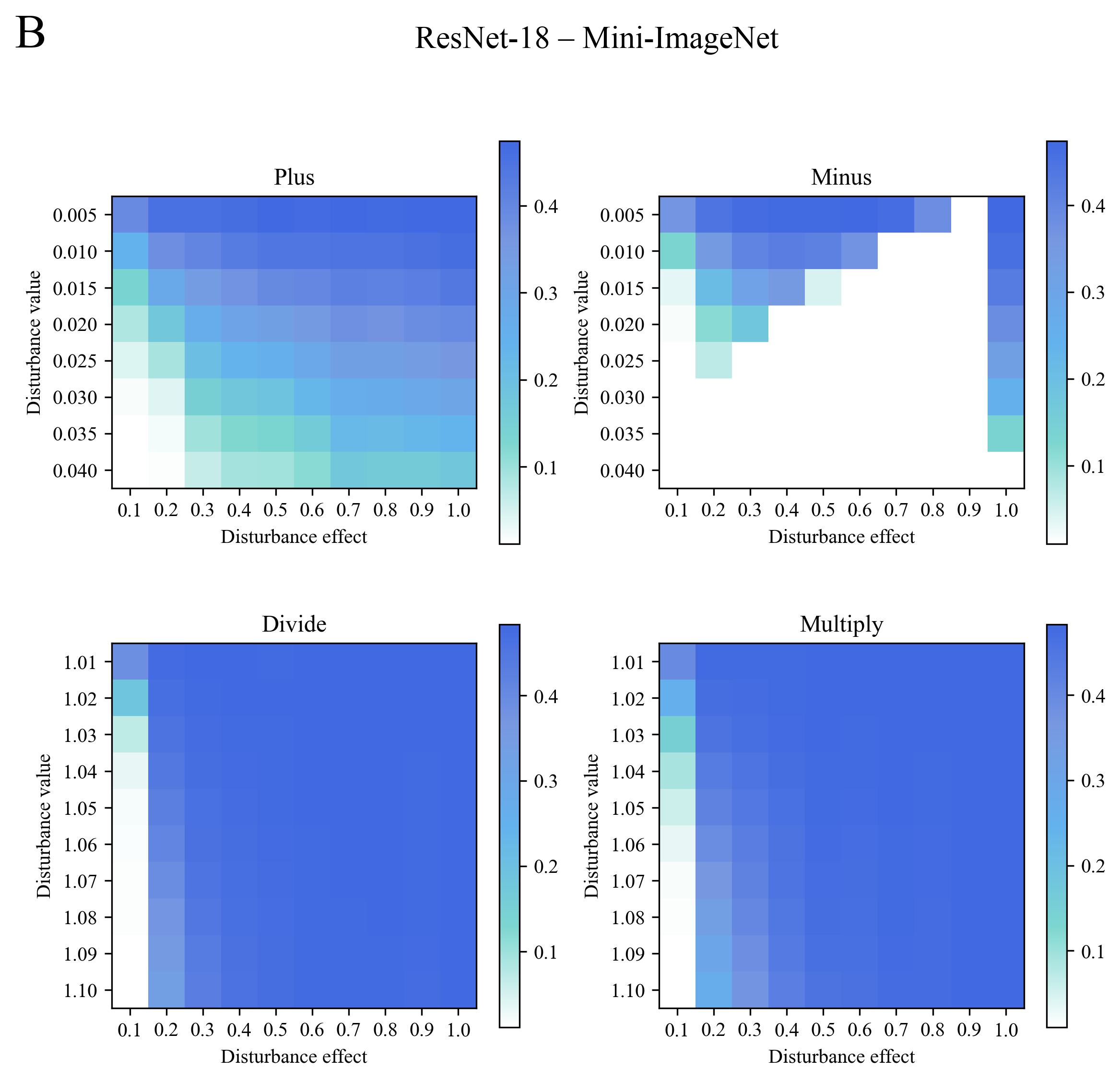}\\[2pt]
\includegraphics[width=0.48\textwidth]{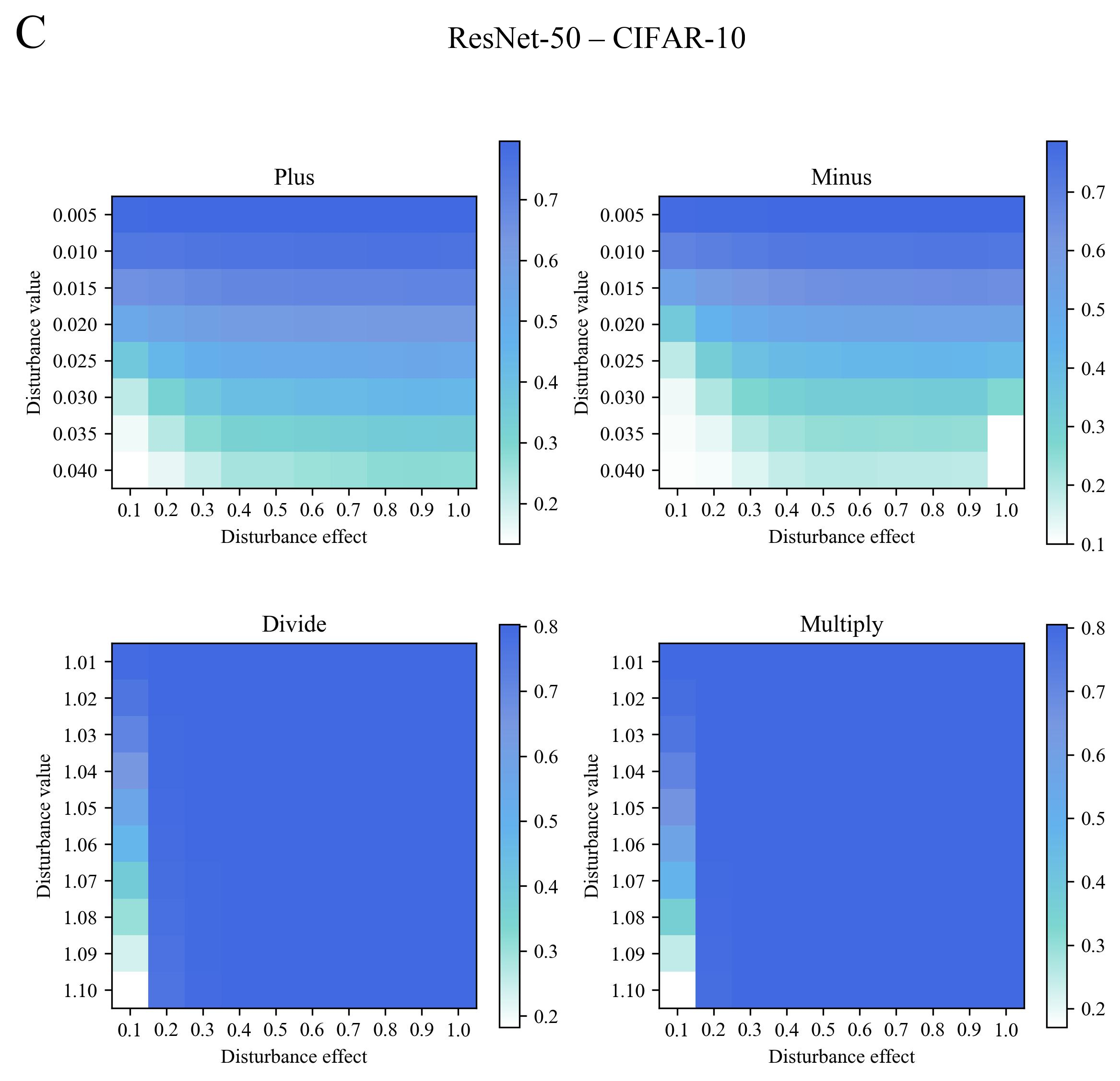}\hfill
\includegraphics[width=0.48\textwidth]{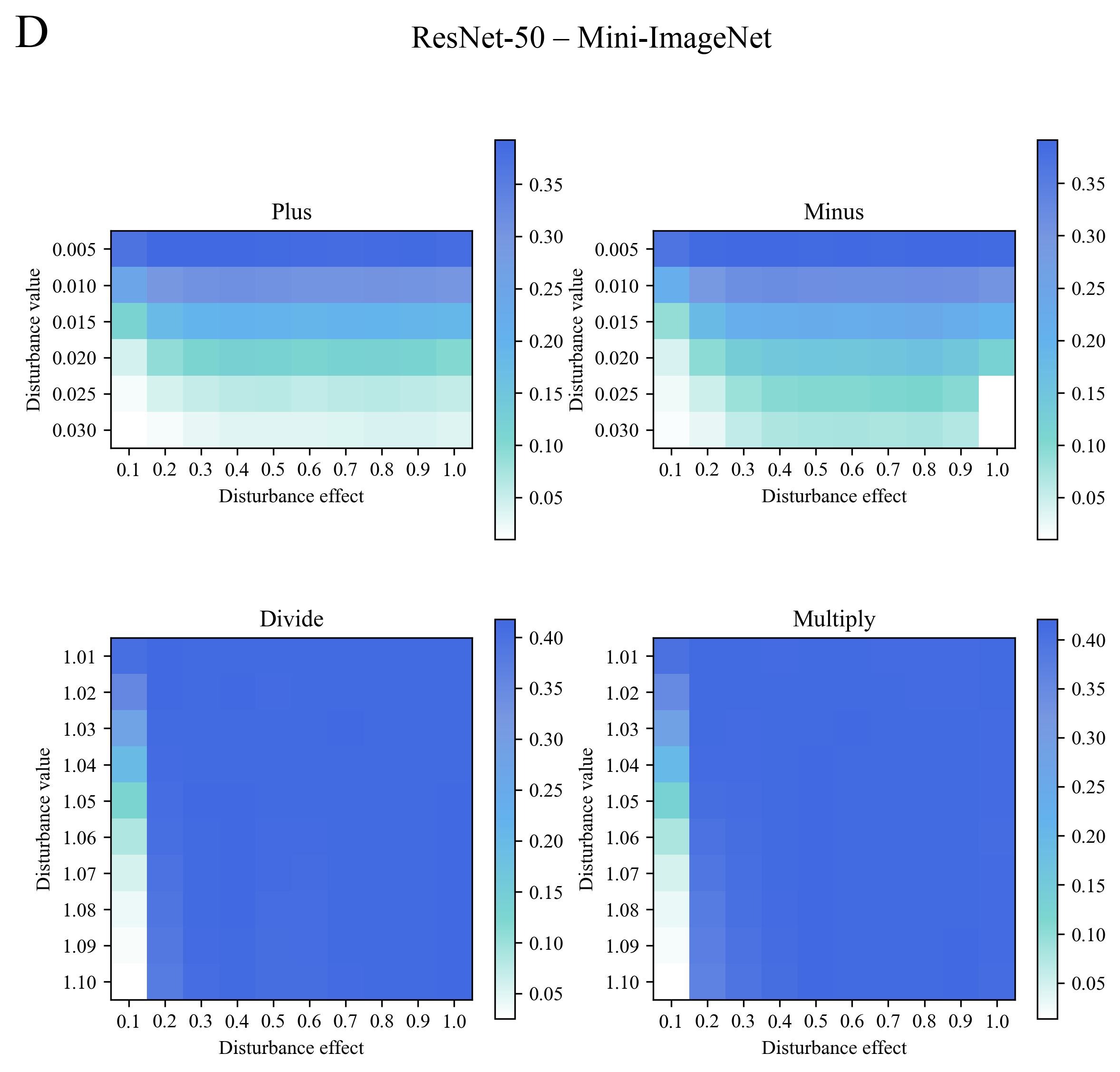}
\caption{Quantitative results of neuron importance across different disturbance intervals for ResNet-18 (A: CIFAR-10, B: Mini-ImageNet) and ResNet-50 (C: CIFAR-10, D: Mini-ImageNet). \emph{Abscissa:} the ten non-overlapping weight intervals, not a second perturbation axis; tick $q$ denotes the interval spanning the $(q-0.1)$--$q$ quantiles of $\lvert w\rvert$ in descending order, so $0.1$ is the top 0--10\% and $1.0$ the bottom 90--100\%. \emph{Ordinate:} the perturbation magnitude $\Delta$. Cell color encodes the absolute post-perturbation test accuracy, darker being higher. Each panel has its own colorbar, with upper limits of approximately 0.8, 0.4, 0.8 and 0.35 for A--D, reflecting the different baseline accuracies attainable on the two datasets; colors are thus comparable within a panel but \emph{not} across panels, and cross-panel comparisons should be read from Tables~\ref{tab:resnet18} and~\ref{tab:resnet50}. ``Plus'' and ``Minus'' add a constant perturbation to the original weights, while ``Multiply'' and ``Divide'' apply a perturbation proportional to the magnitude of the original weights.}
\label{fig:heatmap}
\end{figure*}

\begin{table*}[tp]
\centering
\caption{Perturbation effects on neurons in the ResNet-18 model. Each cell is the \emph{absolute accuracy degradation} (unperturbed test accuracy minus perturbed accuracy, in percentage points); larger is more sensitive. Within each row the magnitude $\Delta$ is fixed at the smallest value producing a measurable degradation in the top 10\% interval and applied unchanged to all ten intervals, so the ten cells are directly comparable. ``Random'' applies the same operation and magnitude to a randomly chosen 10\% of neurons from the whole network and gives the noise floor.}
\label{tab:resnet18}
\resizebox{\textwidth}{!}{%
\begin{tabular}{|c|c|c|c|c|c|c|c|c|c|c|c|c|}
\hline
\multirow{2}{*}{Dataset} & \multirow{2}{*}{Operation} & \multicolumn{10}{|c|}{Disturbance interval (\%)} & \multirow{2}{*}{Random} \\
\cline{3-12}
& & 0--10 & 10--20 & 20--30 & 30--40 & 40--50 & 50--60 & 60--70 & 70--80 & 80--90 & 90--100 & \\ \hline
\multirow{4}{*}{CIFAR-10} & plus & 45.18 & 14.70 & 10.29 & 6.40 & 5.01 & 3.54 & 2.97 & 2.31 & 1.99 & 1.27 & 6.33 \\
& minus & 38.39 & 17.01 & 11.05 & 7.24 & 5.55 & 4.02 & 3.12 & 2.52 & 2.25 & 1.43 & 7.41 \\
& multiply & 74.38 & 5.10 & 2.74 & 1.84 & 2.04 & 1.85 & 1.74 & 1.67 & 1.75 & 1.80 & 5.08 \\
& divide & 79.33 & 3.89 & 2.11 & 1.93 & 1.81 & 1.55 & 1.66 & 1.52 & 1.37 & 1.51 & 3.33 \\ \hline
\multirow{4}{*}{\shortstack{Mini-\\ImageNet}} & plus & 32.86 & 17.60 & 11.05 & 7.83 & 5.47 & 4.85 & 2.93 & 3.27 & 2.39 & 1.07 & 10.66 \\
& minus & 15.69 & 5.78 & 2.77 & 1.96 & 2.39 & 4.46 & 21.37 & 21.37 & 21.37 & 0.47 & 2.36 \\
& multiply & 62.99 & 6.14 & 5.51 & 4.03 & 3.05 & 2.78 & 3.18 & 3.32 & 2.69 & 2.91 & 3.41 \\
& divide & 72.36 & 6.38 & 3.39 & 2.73 & 2.56 & 2.05 & 2.14 & 1.71 & 2.16 & 1.99 & 2.52 \\ \hline
\end{tabular}%
}
\end{table*}

\begin{table*}[tp]
\centering
\caption{Perturbation effects on neurons in the ResNet-50 model, reported as in Table~\ref{tab:resnet18}. All values are percentages (\%).}
\label{tab:resnet50}
\resizebox{\textwidth}{!}{%
\begin{tabular}{|c|c|c|c|c|c|c|c|c|c|c|c|c|}
\hline
\multirow{2}{*}{Dataset} & \multirow{2}{*}{Operation} & \multicolumn{10}{|c|}{Disturbance interval (\%)} & \multirow{2}{*}{Random} \\
\cline{3-12}
& & 0--10 & 10--20 & 20--30 & 30--40 & 40--50 & 50--60 & 60--70 & 70--80 & 80--90 & 90--100 & \\ \hline
\multirow{4}{*}{CIFAR-10} & plus & 33.75 & 18.86 & 11.40 & 6.67 & 6.34 & 4.87 & 3.89 & 2.53 & 1.44 & 2.71 & 7.55 \\
& minus & 39.45 & 18.87 & 11.17 & 7.61 & 4.44 & 3.72 & 2.98 & 1.76 & 1.97 & 3.14 & 4.88 \\
& multiply & 66.63 & 3.77 & 3.04 & 2.79 & 3.18 & 3.49 & 3.54 & 3.35 & 3.38 & 3.49 & 3.35 \\
& divide & 71.94 & 4.95 & 3.06 & 2.57 & 2.34 & 2.32 & 2.12 & 2.21 & 2.23 & 2.34 & 3.93 \\ \hline
\multirow{4}{*}{\shortstack{Mini-\\ImageNet}} & plus & 36.21 & 11.66 & 4.37 & 3.16 & 4.86 & 7.05 & 6.56 & 5.99 & 6.97 & 8.91 & 4.25 \\
& minus & 40.36 & 13.50 & 5.12 & 3.57 & 4.58 & 5.42 & 5.12 & 0.41 & 4.88 & 8.83 & 4.52 \\
& multiply & 63.96 & 3.84 & 3.89 & 3.10 & 2.78 & 3.10 & 3.52 & 3.61 & 3.57 & 3.98 & 4.64 \\
& divide & 55.60 & 4.13 & 4.00 & 4.13 & 5.06 & 4.81 & 3.75 & 4.81 & 4.75 & 4.44 & 4.52 \\ \hline
\end{tabular}%
}
\end{table*}

\subsection{Model Robustness Under Perturbation Strategies}
After partitioning neuron weights into ten intervals of 10\% each, we randomly sampled 10\% of the neurons and conducted ten independent perturbation experiments, using the average as a benchmark. Fig.~\ref{fig:heatmap} reports the perturbation effects across models and datasets, with panels~A and~B for ResNet-18 (CIFAR-10 and Mini-ImageNet) and panels~C and~D for ResNet-50.

In panel~A (ResNet-18, CIFAR-10), high-weight neurons in the top 10\% interval exhibit pronounced vulnerability: under equidistant perturbation, accuracy drops sharply even with only a 5\% weight disturbance, and under proportional perturbation, accuracy declines rapidly as intensity increases, while other intervals show negligible change. In panel~B (ResNet-18, Mini-ImageNet), the top 10\% neurons remain highly vulnerable, and a new trend emerges: under equidistant perturbation (e.g., a fixed disturbance of 0.04), the lowest 10\% weight neurons also experience a steep accuracy drop, indicating threshold-dependent degradation when low-weight perturbations exceed a certain level. Panels~C and~D (ResNet-50) confirm the generalizability of these trends across a deeper architecture: the top 10\% neurons remain vulnerable and low-weight intervals still show threshold-dependent degradation, with only subtle architecture-specific differences in the magnitude of accuracy drops.

Tables~\ref{tab:resnet18} and~\ref{tab:resnet50} provide quantitative support for Fig.~\ref{fig:heatmap}. For each dataset and operation, the perturbation magnitude is fixed at the smallest value that produces a measurable degradation in the top 10\% interval and is then applied unchanged to all ten intervals, so the entries within a row measure the sensitivity of the intervals under an identical perturbation. Both models show that the top 10\% weight neurons have a significantly greater impact on performance than other intervals or random perturbations---for example, under the ``Divide'' operation, the top 10\% neurons in ResNet-18 on CIFAR-10 cause 79.33\% accuracy degradation, far exceeding the 3.33\% from random perturbations. In contrast, low-weight intervals show a weak correlation between weight and importance; some intervals even exhibit improved test accuracy, with no discernible pattern linked to weight distribution.

Two caveats temper the interpretation of these numbers. First, the effect of several mid- and low-weight intervals is comparable to, or smaller than, that of the random baseline---for instance, under ``plus'' on CIFAR-10 the 40\%--100\% intervals all fall below the 6.33\% random value in Table~\ref{tab:resnet18}---so their individual contribution cannot be distinguished from noise without variance estimates, and the evidence that low-weight neurons matter is dataset-dependent rather than conclusive. Second, under proportional perturbation (``multiply''/``divide'') the largest-magnitude weights receive, by construction, the largest absolute change, so their dominant effect partly reflects the perturbation design rather than functional importance alone; the additive operations (``plus''/``minus''), which apply an equal absolute change across intervals, therefore provide the more directly comparable evidence. With these caveats in mind, neuron weight correlates with importance in the high-weight range but shows weak or inconsistent links in lower ranges, highlighting the complexity of the weight--importance relationship.

\subsection{Performance After Removing High-Weight Neurons and Retraining}
\begin{figure*}[tp]
\centering
\includegraphics[width=0.9\textwidth]{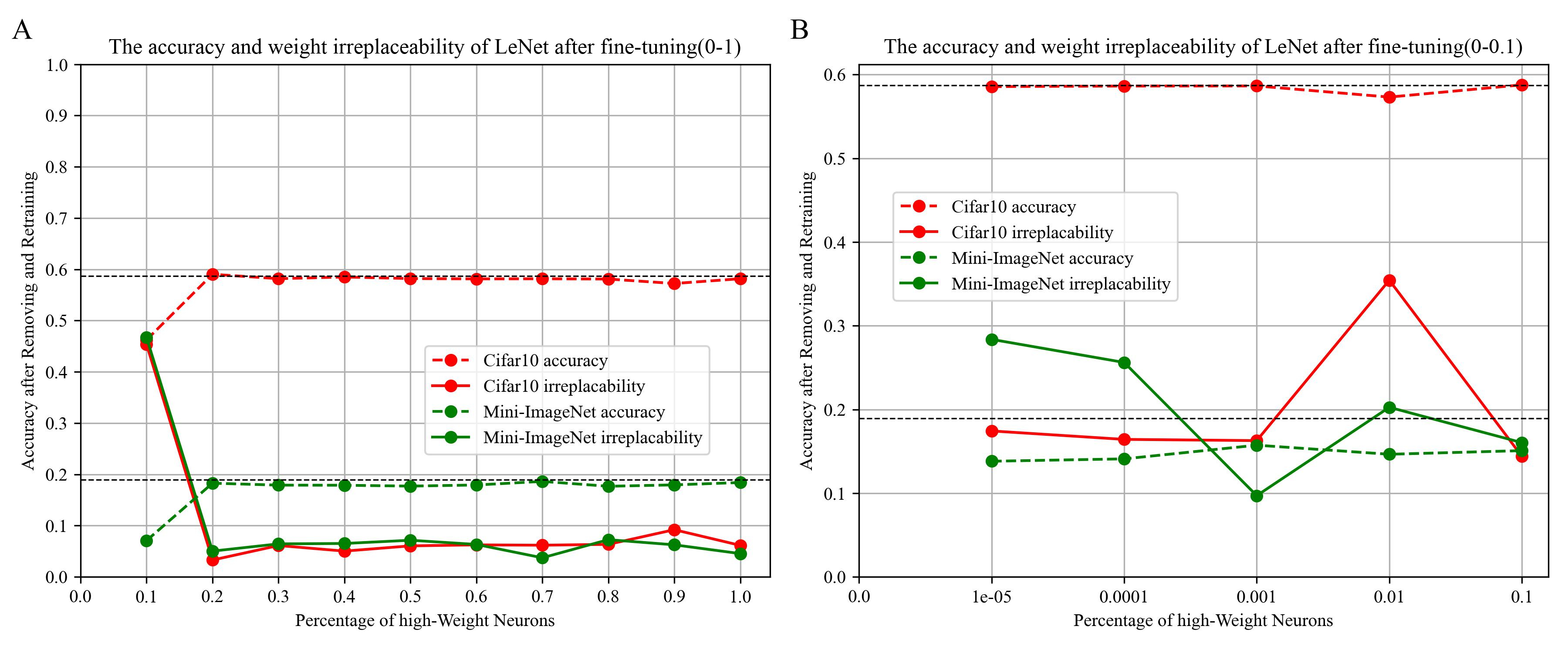}
\includegraphics[width=0.9\textwidth]{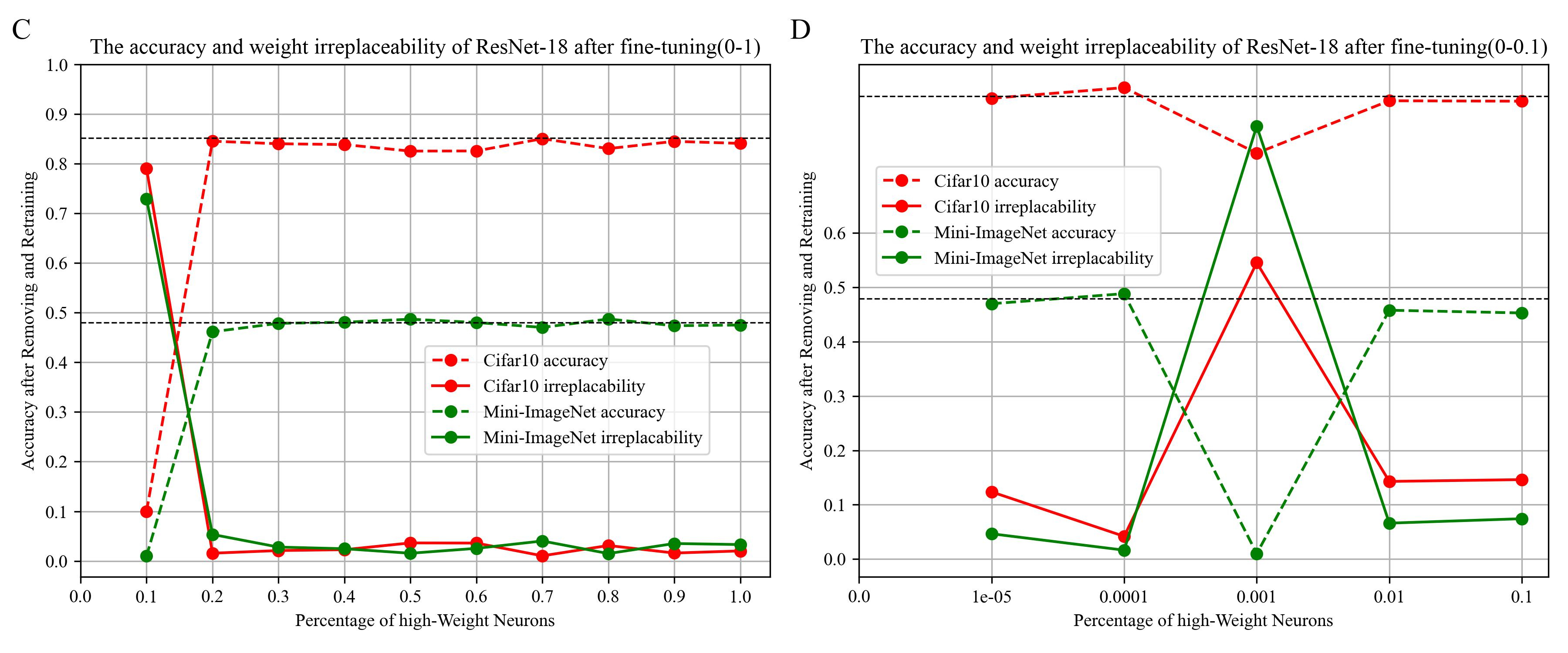}
\includegraphics[width=0.9\textwidth]{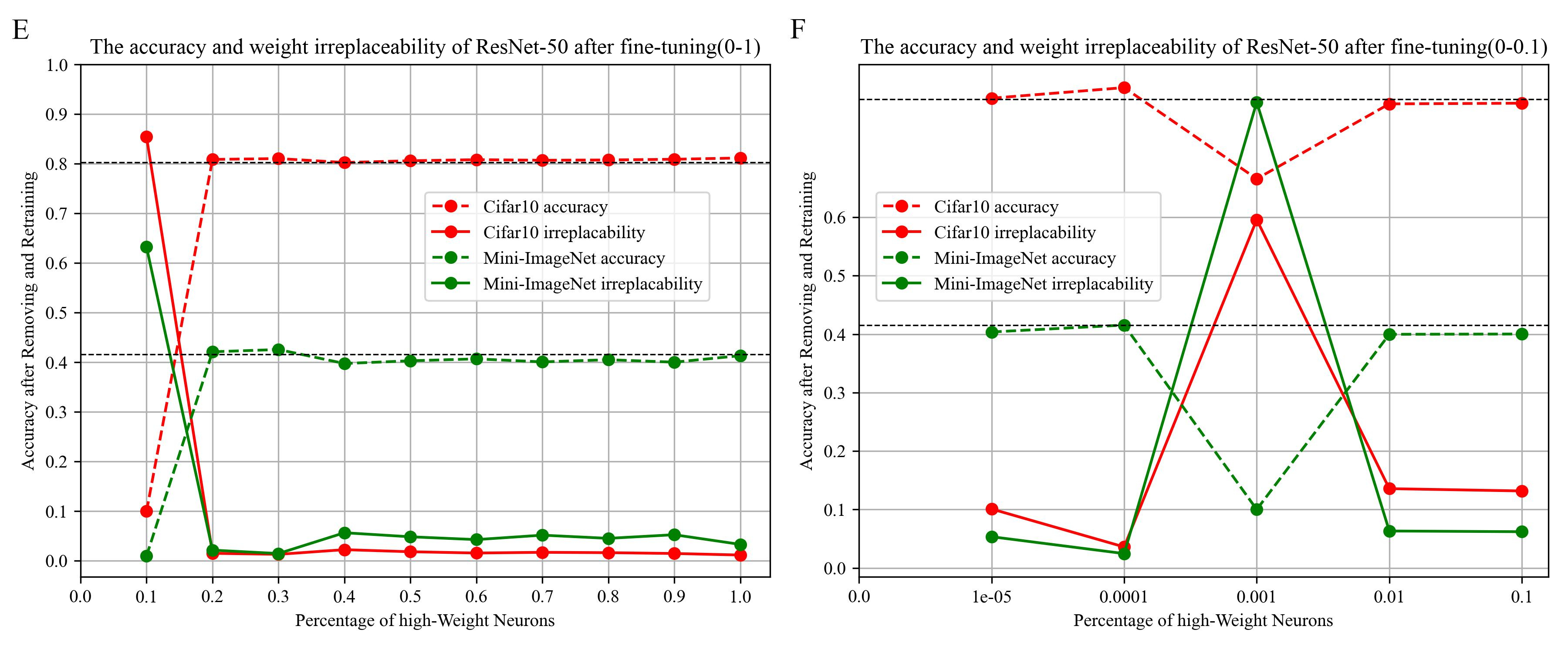}
\caption{Neuron removal and retraining results for LeNet (panels A, B), ResNet-18 (panels C, D), and ResNet-50 (panels E, F) on CIFAR-10 and Mini-ImageNet. Red curves correspond to CIFAR-10 and green curves to Mini-ImageNet; dashed curves denote the post-retraining test accuracy and solid curves denote the neuron irreplaceability. The black dashed line indicates the original test accuracy. Panels A, C, and E scan the full range of 10\% intervals, whereas panels B, D, and F zoom into the top-10\% region, subdividing it into finer sub-intervals ($10^{-5}$ to $0.1$).}
\label{fig:retrain}
\end{figure*}

On CIFAR-10 and Mini-ImageNet, we divided the weights of LeNet, ResNet-18, and ResNet-50 into ten equal intervals, applied neuron ablation to each, and fine-tuned the models; ten control experiments with random neuron ablation ensured the validity of the results. Fig.~\ref{fig:retrain} reports these results, with panels~A and~B for LeNet, panels~C and~D for ResNet-18, and panels~E and~F for ResNet-50.

The full-range panels (A, C, E) reveal a clear pattern: removing the top 10\% absolute-weight neurons leaves the model's performance stuck at the random-ablation baseline even after 20 epochs of retraining, and the neuron irreplaceability peaks accordingly. For ResNet-18 and ResNet-50 on CIFAR-10, accuracy collapses to roughly chance level (about 10\%) after top 10\% ablation and does not recover, whereas ablating any other weight interval allows the model to reallocate weights during retraining, recovering test accuracy and sometimes exceeding the original baseline. LeNet (panel~A) shows a distinct but related trend: its performance drops markedly after top 10\% ablation but stays above the random-guessing level, indicating architectural differences in weight sensitivity. This holds consistently across both datasets (red for CIFAR-10, green for Mini-ImageNet), ruling out dataset-specific bias and confirming that the irreplaceability of the top 10\% neurons generalizes across architectures and data distributions.

Subdividing the top 10\% into finer sub-intervals (panels B, D, F; from $10^{-5}$ to $0.1$) reveals a more nuanced picture. For ResNet-18 and ResNet-50, the irrecoverable damage is localized to a narrow sub-interval around the top 0.1\%: ablating this band causes a spike in irreplaceability and a dip in post-retraining accuracy, whereas ablating the other high-weight sub-intervals permits near-full recovery through fine-tuning. Collectively, Fig.~\ref{fig:retrain} demonstrates that the top 10\% absolute-weight neurons are uniquely irreplaceable across models and datasets, that this irreplaceability concentrates in a very narrow high-weight band, and that other weight intervals can be compensated for via retraining.

\subsection{Discussion}
This study uses neuron ablation, perturbation, and ablation--retraining experiments to quantify the non-trivial relationship between neuron weights and their importance, revealing nuanced roles of high-weight neurons and critical contributions from low-weight neuron collectives.

The overlap analysis shows that top-weight neurons dominate individual importance, but lower-weight intervals exhibit an increasing collective contribution. Subdividing high-overlap intervals reveals an importance gradient even among relatively high-weight neurons, suggesting that low-weight neurons encode widespread feature combinations vital to model functionality, while high-weight neurons vary in task-relevant feature encoding.

The perturbation testing demonstrates that the top 10\% weight neurons are highly sensitive---small perturbations drastically degrade performance by disrupting key recognition features. In some intervals and datasets, low-weight neurons also affect performance more than mid-range ones when perturbed, possibly due to their role in bridging essential feature combinations; however, as noted above, these low-weight effects are close to the random baseline and warrant confirmation with variance estimates before strong conclusions are drawn.

The ablation--retraining experiments find that removing the broad top 10\% high-weight neurons causes irreversible performance loss, but the difficulty is concentrated in a narrow top-0.1\% band; ablating the remaining high-weight sub-intervals allows full recovery via fine-tuning. This indicates that these extreme high-weight neurons occupy local critical nodes in information pathways, yet the model can reconstruct effective processing through other neurons' weight adjustments.

Overall, this work does not support the monotonic weight--importance assumption. Two results hold consistently across architectures and datasets and carry the argument: the overlap between the high-weight set and the accuracy-critical set is uniformly low, and irrecoverable damage is confined to a very narrow band at the top of the weight ranking rather than spread over the high-weight region as a whole. The non-monotonic responses observed in some low-weight intervals point in the same direction, but they appear only for particular dataset--operation combinations and remain close to the random baseline, so we treat them as suggestive rather than as evidence that low-weight neurons are indispensable. In addition to weight magnitude, a neuron's contribution to the output is also influenced by factors such as its position within the network, the pattern of inter-neuronal connections, the initial weight values, and the training process.

Several methodological limitations should be kept in mind when interpreting these results. (i)~As noted in Section~\ref{sec:method}, importance is measured at the granularity of individual weight parameters rather than whole units; setting a single parameter to zero is not equivalent to ablating a conventional neuron, so our conclusions concern weight magnitude rather than unit-level function. (ii)~Single-parameter ablation produces near-noise-level accuracy changes in large networks, making the accuracy-impact ranking $S_a$---and hence the overlap analysis---most reliable on small models such as LeNet and noisier on ResNet. (iii)~In the ablation--retraining experiment, zeroing and freezing the largest-magnitude weights removes a substantial fraction of the network's representational capacity, so the failure to recover after top-10\% ablation may partly reflect capacity loss rather than the specific irreplaceability of those units. (iv)~Results are reported without variance estimates or significance tests, and the base accuracies (particularly on Mini-ImageNet) are modest, so the quantitative magnitudes should be read as indicative. In the future, we aim to address these points with unit-level ablation, proper chance and significance baselines, and stronger backbones, and to extend the study to domains such as object detection and natural language processing with a more granular, module- or layer-level analysis.

\section{Conclusion}\label{sec:conclusion}
This study presents three strategies to quantify the relationship between neuron weights and their importance: overlap analysis to examine the overlap between high-weight and highly important neurons; perturbation experiments to measure performance changes after neuron perturbations; and ablation experiments that remove high-weight neurons, retrain the model, and assess the impact on the remaining performance. While high-weight neurons play a key role in enhancing robustness and maintaining high accuracy relative to other weight intervals, their importance is not monotonically consistent: a substantial fraction of high-weight neurons are not critical to overall performance, and the accuracy-critical set overlaps only weakly with the high-weight set. This work provides a new perspective and quantitative method for understanding the weight--importance relationship, enriching the theoretical foundation of neural network analysis. The results also suggest, but do not establish, several practical directions: importance-aware, hierarchical protection or pruning schemes---for example, prioritizing critical high-weight parameters during encryption or removing non-critical ones during pruning---and, more speculatively, defenses against pre-training weight contamination, backdoor implantation, and parameter tampering. None of these applications was implemented or evaluated here; the security-oriented ones in particular rest on the untested premise that an attacker's leverage tracks the weight-magnitude ranking, and we regard their empirical verification as future work.

\section*{Acknowledgement}
This work was supported in part by the Shenzhen Science and Technology Program under Grant KJZD20240903104400001, and in part by the Research Task Assignment Project of the Guangdong Laboratory of Artificial Intelligence and Digital Economy (SZ) under Grant GML-26420004.

\bibliographystyle{IEEEtran}
\bibliography{references}

\appendix

\section{Generalization Experiment}\label{secA1}

\begin{figure}[htbp]
  \centering
  \includegraphics[width=0.9\columnwidth]{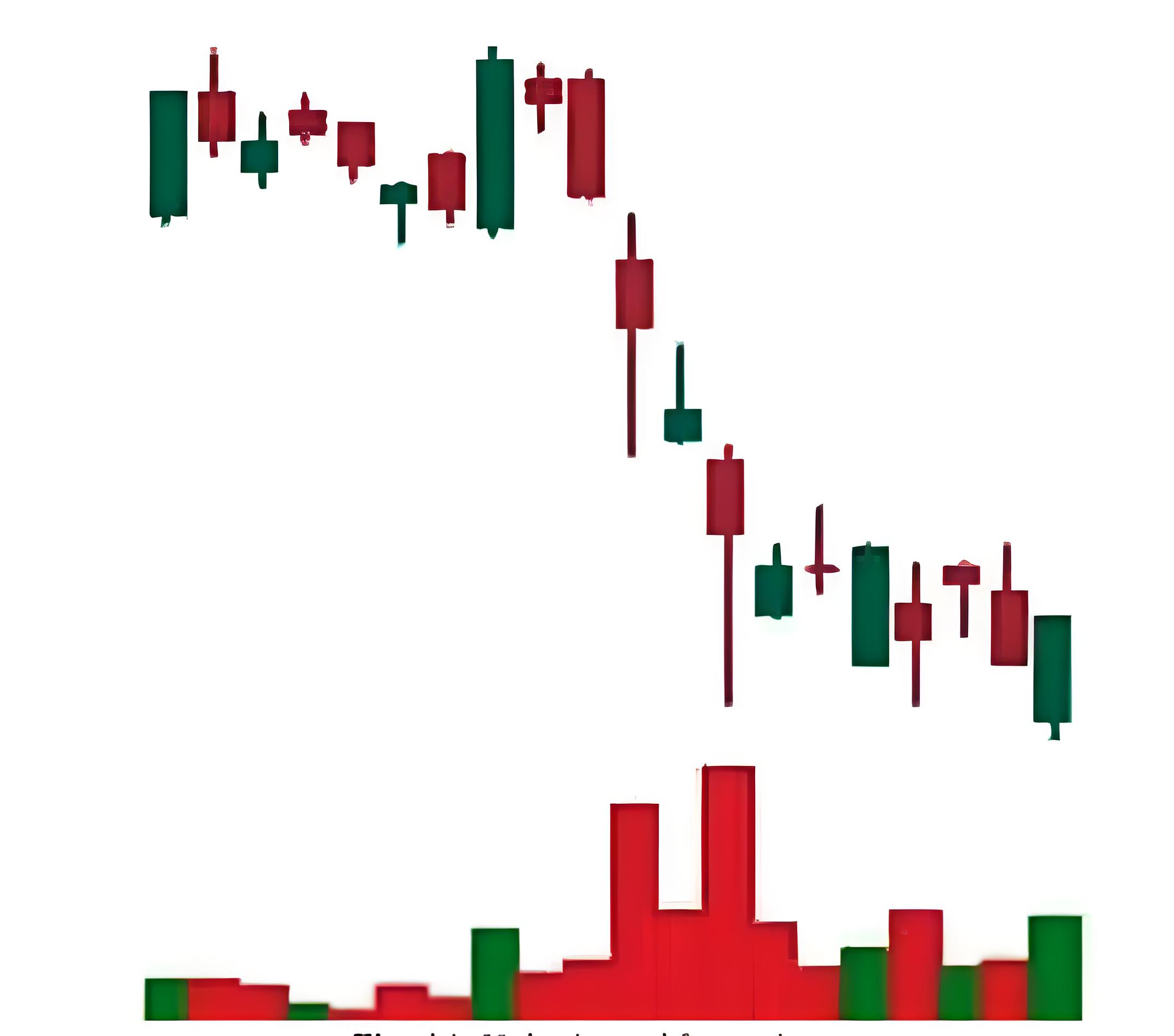}
  \caption{20-day interval futures images. OHLC bar charts are used as the core representation, where the highest and lowest prices are clearly marked by the top and bottom of the bars, while the opening and closing prices are accurately displayed by short horizontal lines on the left and right sides of the bars. A visual element of daily trading volume is added at the bottom of the chart to fully reflect market dynamics.}
\end{figure}

\begin{table}[htbp]
  \centering
  \caption{The Overlap Between High-Weight Neurons and Key Accuracy Contributors}
  \begin{tabular}{ccc}
    \toprule
    \% of High-weight Neurons & Accuracy & Overlap Ratio \\
    \midrule
    0--10   & 54.75 & 42 \\
    10--20  & 54.66 & 19 \\
    20--30  & 54.50 & 16 \\
    30--40  & 54.38 & 12 \\
    40--50  & 54.32 &  6 \\
    50--60  & 54.28 &  4 \\
    60--70  & 54.20 &  8 \\
    70--80  & 54.11 &  9 \\
    80--90  & 54.08 &  3 \\
    90--100 & 53.98 & 10 \\
    \bottomrule
  \end{tabular}
  \label{tab:a1}
\end{table}

\begin{table}[htbp]
  \centering
  \caption{The Model Robustness Under Perturbation Strategies}
  \resizebox{\columnwidth}{!}{%
  \begin{tabular}{cccccccccccc}
    \toprule
    \multirow{2}{*}{Operation} & \multicolumn{10}{c}{Disturbance interval (\%)} & \multirow{2}{*}{Random} \\
    \cmidrule(lr){2-11}
    & 0--10 & 10--20 & 20--30 & 30--40 & 40--50 & 50--60 & 60--70 & 70--80 & 80--90 & 90--100 & \\
    \midrule
    plus     & 35.71 & 14.32 &  9.87 & 6.18 & 5.14 & 3.72 & 3.04 & 2.45 & 2.11 & 1.39 & 5.07 \\
    minus    & 23.81 & 16.74 & 10.89 & 7.12 & 5.47 & 4.08 & 3.21 & 2.63 & 2.37 & 1.52 & 7.16 \\
    multiply & 73.53 &  5.07 &  2.79 & 1.87 & 2.09 & 1.89 & 1.78 & 1.69 & 1.73 & 1.84 & 5.72 \\
    divide   & 78.74 &  3.88 &  2.13 & 1.92 & 1.83 & 1.59 & 1.64 & 1.51 & 1.40 & 1.43 & 4.19 \\
    \bottomrule
  \end{tabular}}
  \label{tab:a2}
\end{table}

\begin{table}[htbp]
  \centering
  \caption{Performance Evaluation After Removing High-Weight Neurons and Retraining}
  \begin{tabular}{ccc}
    \toprule
    \% of High-weight Neurons & Accuracy & Irreplaceability \\
    \midrule
    0--10   & 50.00 & 56.75 \\
    10--20  & 54.62 &  6.04 \\
    20--30  & 54.77 &  5.49 \\
    30--40  & 54.85 &  4.61 \\
    40--50  & 54.81 &  5.05 \\
    50--60  & 55.02 &  6.04 \\
    60--70  & 55.03 &  3.73 \\
    70--80  & 54.94 &  3.62 \\
    80--90  & 54.86 &  4.50 \\
    90--100 & 54.89 &  4.17 \\
    \bottomrule
  \end{tabular}
  \label{tab:a3}
\end{table}

\subsection{Generalization Experiment Setup}

In addition to open-source data, to further validate our conclusions, we collected real financial data to construct our own dataset and conducted experiments on this dataset. The specific implementations are as follows.

Through the JoinQuant quantitative investment and research platform, we collected 50 futures contracts from major domestic trading markets such as the China Financial Futures Exchange, Shanghai Futures Exchange, and Dalian Commodity Exchange, fully covering historical daily data from 2003 to 2024.

Given the inherent time constraints of futures contracts --- that is, the limited contract holding period and the mandatory delivery at maturity --- there are significant overlapping characteristics among different futures contracts in the temporal dimension. To ensure the continuity and completeness of time-series analysis, the preprocessing of futures contracts is particularly important. The strategy adopted in this study is as follows: first, the acquired futures contract data are spliced in chronological order, and the contract with the largest daily trading volume is selected as the main contract for that day, serving as the primary representative of market trends. Subsequently, to eliminate abnormal price fluctuations caused by irregular contract changes, this study introduces median filtering technology as a preliminary smoothing method, then plots the contract month change diagram through MATLAB, followed by manual review and adjustment, especially at the switching points of main contracts, to ensure data continuity and accuracy.

In terms of data visualization, this study follows the graphic design principles of industry-standard websites such as Bloomberg and Yahoo Finance to construct major price charts (Fig.~A1). The image sequence designed in this study is composed of OHLC bar charts at 20-day intervals (monthly strategy). By setting a unified image height and dynamically adjusting the vertical axis ratio, the highest and lowest points of the OHLC path in each image are precisely aligned with the top and bottom edges of the image. This design strategy ensures that the price ranges of all securities are uniformly scaled to the same height, eliminating differences in absolute prices across securities, allowing the model to focus more on relative price changes, and enhancing the consistency of cross-asset predictions.

\subsection{Generalization Experiment Result}

LeNet architecture was employed for training, with training results compared against CIFAR-10 and Mini-ImageNet datasets. Financial data exhibits unique characteristics due to its strong temporal correlation and complex noise interference. Nevertheless, universal patterns in the importance distribution of neurons across cross-domain models can still be identified from such data.

We flattened the weight matrix of the LeNet model trained on the futures price trend image dataset into a one-dimensional matrix, took the absolute values of its elements, sorted them by magnitude, and then conducted individual neuron ablation experiments. As shown in Table~\ref{tab:a1}, the trend features in financial images (such as ``Evening Star'' and ``Golden Cross'') rely on a small number of high-weight neurons for strong semantic encoding. Meanwhile, neurons with weights below 10\% have a higher overlap rate compared to those with weights between 40\% and 90\%. This phenomenon indicates that there is no monotonically decreasing relationship between the weight of a neuron and its importance.

The preprocessed neuron matrix was split according to weight intervals, followed by non-ablation perturbation experiments. As shown in Table~\ref{tab:a2}, we still observed a trend where the model sensitivity is inversely proportional to the decrease in weights in the non-ablation perturbation experiments. This indicates that low-weight neurons can exert a greater impact on model performance compared to neurons with medium weights.

The output neuron matrix was split by weight, and each interval was subjected to ablation, followed by fine-tuning with both the gradient and the sum of gradients maintained at zero during the fine-tuning process. As shown in Table~\ref{tab:a3}, after removing neurons in each interval, the model performance can be partially restored through fine-tuning. This indicates that although the individual contribution of neurons in low-weight intervals (e.g., 90--100\%) is limited, their collective collaboration through complex feature combinations significantly affects the generalization ability of the model.

This experiment also reveals the same law in the distribution of neuron weights in financial image models: a few high-weight neurons serve as the ``engines'' of decision-making but are vulnerable to perturbations; a large number of low-weight neurons form an ``ecosystem'' that implicitly supports model generalization.

\end{document}